\documentclass[journal]{IEEEtran}
\usepackage{times}
\usepackage{epsfig}
\usepackage{graphicx}
\usepackage{amsmath}
\usepackage{amssymb}
\usepackage{xcolor}
\usepackage{relsize}
\usepackage[ruled]{algorithm2e}
\usepackage{soul}

\ifCLASSINFOpdf
\else
\fi
\SetKwRepeat{Do}{do}{while}
\begin{document}
	%
	\title{Image Co-skeletonization \emph{via} Co-segmentation}
	%
	%
	%
	
	\author{Koteswar~Rao~Jerripothula,~\IEEEmembership{Member,~IEEE,}
		Jianfei~Cai,~\IEEEmembership{Senior Member,~IEEE,}~Jiangbo~Lu,~\IEEEmembership{Senior Member,~IEEE,}
		and~Junsong~Yuan,~\IEEEmembership{Senior Member,~IEEE}\thanks{A preliminary version of this work appears in CVPR'17 \cite{8099896}. K. R. Jerripothula is with IIIT Delhi, India, J. Cai is with Monash University, Australia, J. Lu is with Shenzhen Cloudream Technology Co., Ltd. and J. Yuan is with University at Buffalo, United States of America. E-mail: krjimp@geu.ac.in, asjfcai@ntu.edu.sg, jiangbo.lu@gmail.com and jsyuan@buffalo.edu. Contact krjimp@geu.ac.in for further questions about this work.
		}
	}
	
	%
	%

	\markboth{Submitted to IEEE Transactions on Image Processing}%
	{Shell \MakeLowercase{\textit{et al.}}: Bare Demo of IEEEtran.cls for IEEE Journals}
	%



	\maketitle
	
    \begin{abstract}
        Recent advances in the joint processing of images have certainly shown its advantages over individual processing. Different from the existing works geared towards co-segmentation or co-localization, in this paper, we explore a new joint processing topic: image co-skeletonization, which is defined as joint skeleton extraction of objects in an image collection. Object skeletonization in a single natural image is a challenging problem because there is hardly any prior knowledge about the object. Therefore, we resort to the idea of object co-skeletonization, hoping that the commonness prior that exists across the images may help, just as it does for other joint processing problems such as co-segmentation. We observe that the skeleton can provide good scribbles for segmentation, and skeletonization, in turn, needs good segmentation. Therefore, we propose a coupled framework for co-skeletonization and co-segmentation tasks so that they are well informed by each other, and benefit each other synergistically. Since it is a new problem, we also construct a benchmark dataset by annotating nearly 1.8k images spread across 38 categories. Extensive experiments demonstrate that the proposed method achieves promising results in all the three possible scenarios of joint-processing: weakly-supervised, supervised, and unsupervised. 
    \end{abstract}
    

    %
    \IEEEpeerreviewmaketitle

    \section{Introduction}
    
    Our main objective in this paper is to exploit joint processing \cite{rother2006cosegmentation,7997806,Jerripothula2016,7919200,Zhu201612,7457899, Meng201667,8290832, 6751215, 5995415,jerripothula2017co} to extract skeletons of the objects in natural images. We call it \emph{object co-skeletonization}. By objects, we mean something which interests the viewer more compared to the background regions such as sky, roads, mountains, and sea in its presence. Automatic skeletonization of such objects has many applications such as image search, image synthesis, and training data generation for object detectors~\cite{Yu:2018:RDD:3304415.3304573}. However, it is difficult to solve this problem as a standalone task, for it requires some support or other. In literature, existing methods either need pre-segmentation~\cite{Choi2003721, Shen2013} of the object in the image or ground-truth skeletons for the training images to learn~\cite{Tsogkas2012, Shen2016306} to perform skeletonization on test images. Even the recent deep learning \cite{8700608,7775034,7529190} based method~\cite{Shen_2016_CVPR} (extended in \cite{shen2017deepskeleton}) requires not only the skeleton location information for training but also the skeleton scale information. Such scale information essentially accounts for shape information, since it is the distance between a skeleton point and the nearest boundary point of the object. Providing such additional scale annotation has shown significant improvement in performance for \cite{Shen_2016_CVPR} over \cite{7410521} that can make use of only skeleton annotations while taking the deep learning \cite{8300630, chen2014semantic, he2017mask} approach.
    
\begin{figure}[t]
        \begin{center}
            \includegraphics[width=1\linewidth]{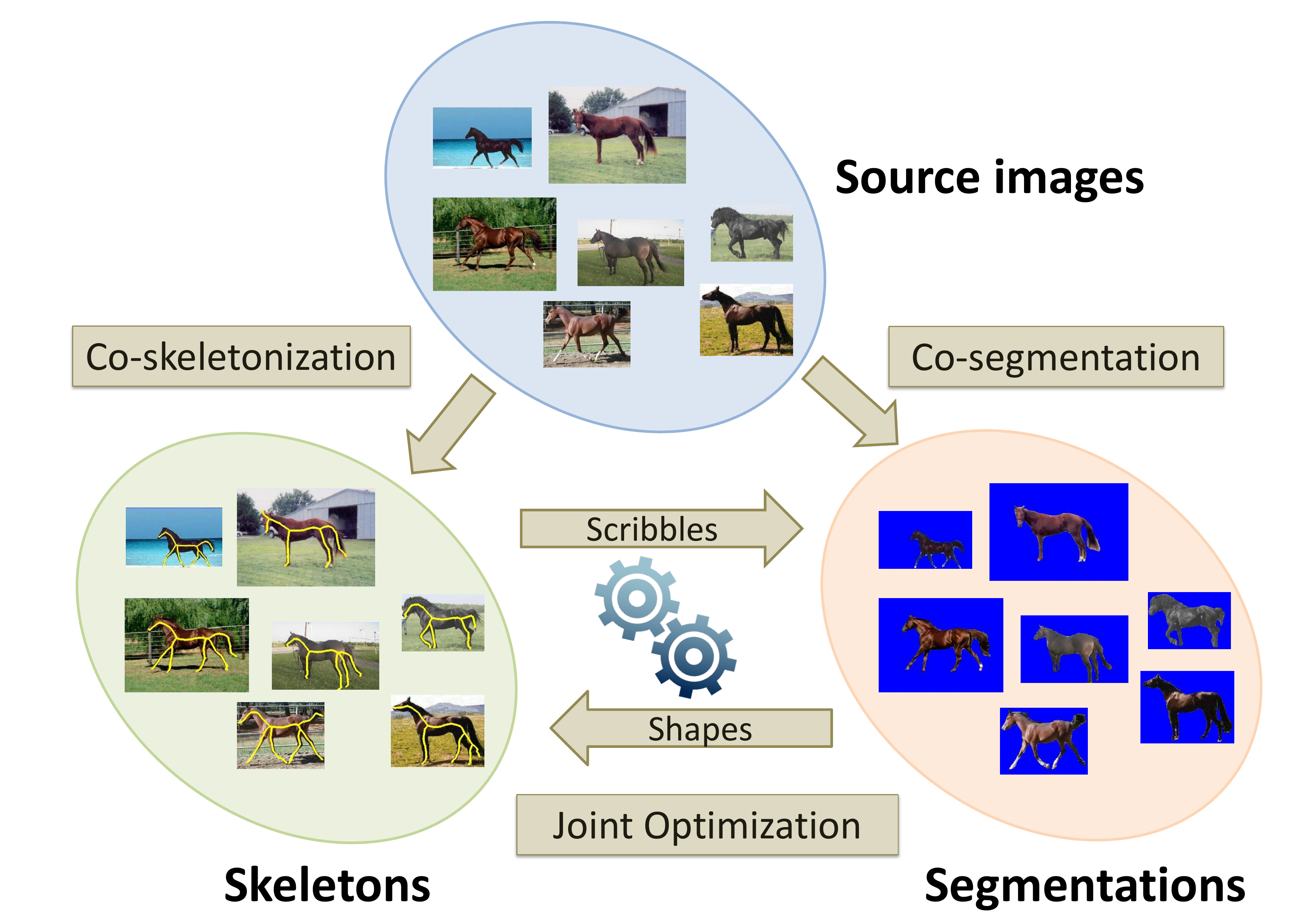}
        \end{center}
        \vspace{-1mm}
        \caption{Image co-skeletonization \emph{via} co-segmentation. Skeletons are in yellow.}\label{fig:jointop}
    \end{figure}

    In contrast, in this paper, we propose a problem called co-skeletonization, which doesn't need to depend upon any kind of annotations exclusively. Instead, we rely on the joint processing, where the central idea is to exploit the commonness of the images. Also, inspired by \cite{Shen_2016_CVPR}, we leverage already existing another joint processing idea named object co-segmentation \cite{rother2006cosegmentation,7583735,Jerr1410:Automatic,8269367} to provide the additional required shape information to our co-skeletonization problem. Interestingly, it turns out that co-skeletonization can also help co-segmentation in return by providing scribble information required for segmentation. In this way, both co-skeletonization and co-segmentation benefit each other synergistically. We couple these two tasks to achieve what we call "\textit{Image Co-skeletonization \emph{via} Co-segmentation}," as shown in Fig.~\ref{fig:jointop}. Unlike the existing methods like \cite{Shen_2016_CVPR,7410521}, which function only in the supervised scenario, by exploiting commonness, the proposed method manages to function in all the three possible scenarios: weakly-supervised, supervised, and unsupervised.   
    
\begin{figure}
        \begin{center}
            \includegraphics[width=1\linewidth]{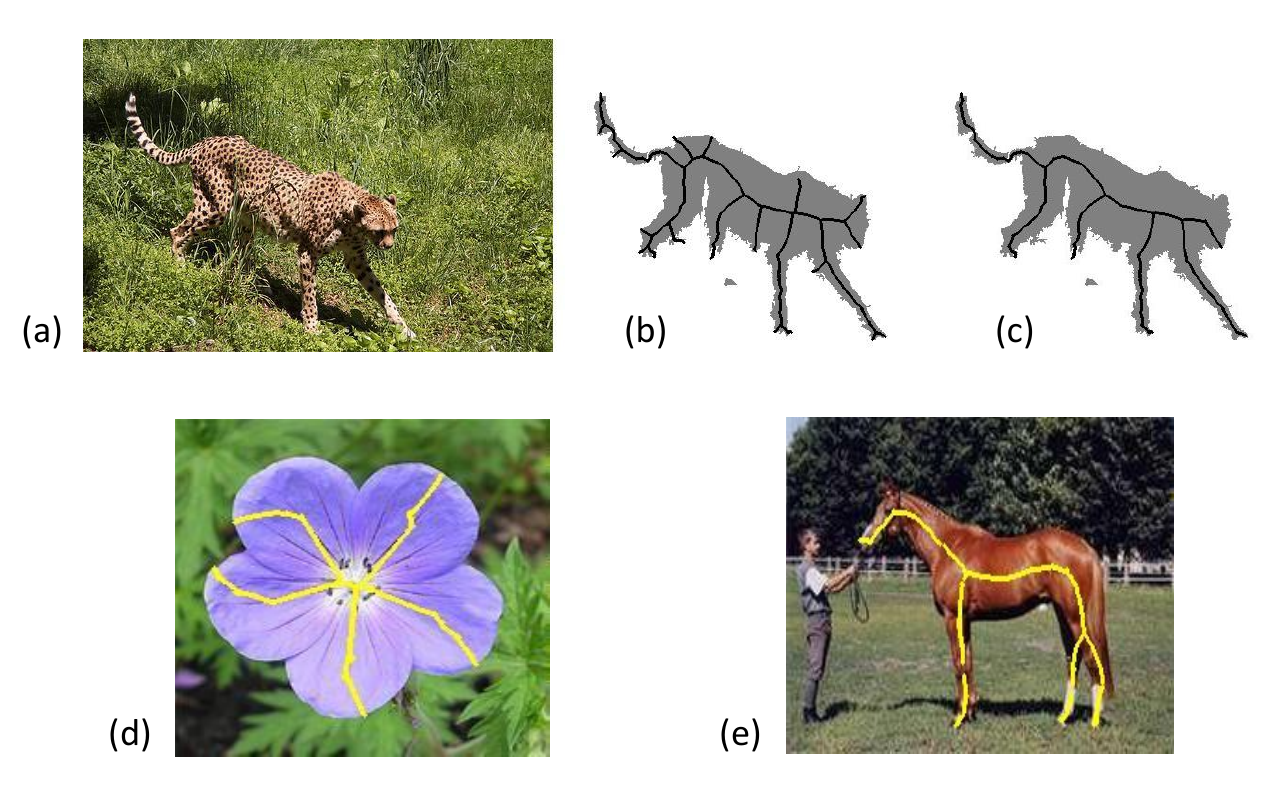}
        \end{center}
        \vspace{-1mm}
        \caption{Example challenges of co-skeletonization. The quality of segmentation affects the quality of skeletonization. (b) The result of \cite{Shen2013} for (a). (c) Our result. Skeletons lie on homogeneous regions, such as in (d) and (e), which are difficult to be detected and described.} \label{fig:chal}
        
    \end{figure}

    There are several challenges involved in performing co-skeletonization and coupling it with co-segmentation. First, existing skeletonization algorithms~\cite{Shen2013, Saha20163, Choi2003721, Shen2011196} can yield a good skeleton provided a good and smooth shape is provided, for they are quite sensitive to the given shape, as shown for the image in Fig.~\ref{fig:chal}(a) which has unsmooth segmentation. The skeleton produced by \cite{Shen2013} in Fig.~\ref{fig:chal}(a) has too many unnecessary branches, while a more desirable skeleton to represent the cheetah would be the one obtained by our modified method in Fig.~\ref{fig:chal}(c). Thus, the quality of the provided shape becomes crucial, which is challenging for the conventional co-segmentation methods because their complicated way of co-labeling many images may not provide good and smooth shapes. Second, the joint processing of skeletons across multiple images is quite tricky. Because most of the skeleton points generally lie in homogeneous regions, as shown in Fig.~\ref{fig:chal}(d) and (e), it is not easy to detect and describe them for matching. Third, how to couple the two tasks so that they can synergistically assist each other is another challenge.

\begin{figure}[t]
        \begin{center}
            \includegraphics[width=1\linewidth]{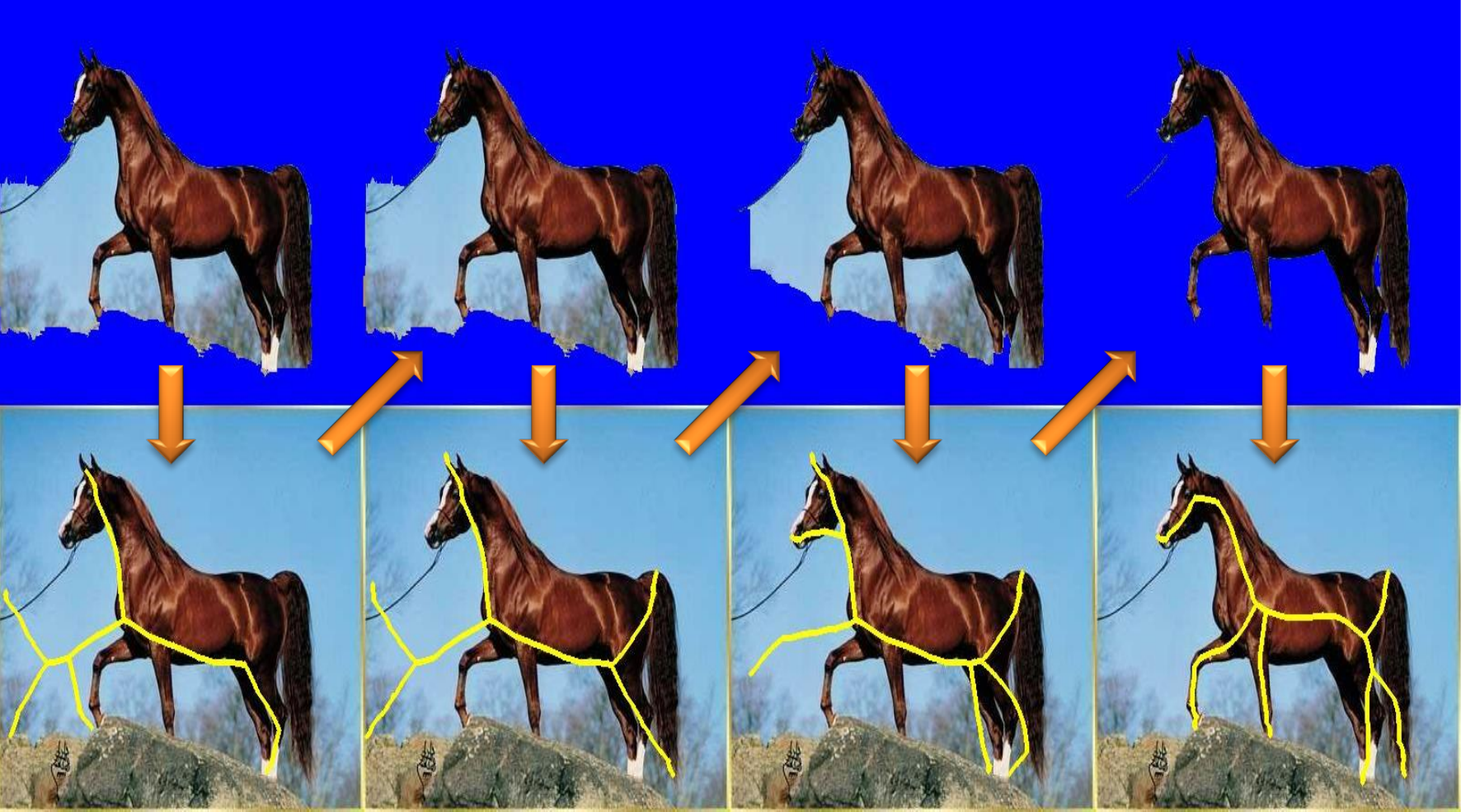}
        \end{center}
        \vspace{-1mm}
        \caption{Inherent interdependencies of co-skeletonization and co-segmentation can be exploited to achieve better results through a coupled iterative optimization process.}
        \label{fig:mot}
        
    \end{figure}

    Our key observation is that we can exploit the inherent interdependencies of the two tasks to achieve better results jointly. For example, in Fig.~\ref{fig:mot}, although the initial segmentation is poor, most of the skeleton pixels remain on the horse body. These skeleton pixels gradually improve the segmentation by providing good scribble information for segmentation in the subsequent iterations of joint processing. In turn, skeletonization also becomes better as the segmentation improves. Our other observation is that we can exploit the structure-preserving property of dense correspondence techniques to overcome the skeleton matching problem. Thanks to the smoothness constraint in its optimization framework, dense correspondence has this useful structure-preserving property.        
    
    
    To the best of our knowledge, there is only one dataset where co-skeletonization could be performed in a weakly-supervised manner (with similar images collected at one place), i.e., WH-SYMMAX dataset~\cite{Shen2016306}, and it only contains horse images. To extensively evaluate co-skeletonization on several categories, we constructed a benchmark dataset named CO-SKEL dataset in our preliminary work \cite{8099896}. It consisted of images ranging from animals, birds, flowers to humans with a total of 26 categories. However, the size of the dataset was small (about 350 images only). In this paper, we present a larger and more challenging dataset containing around 1.8k images, which are categorized into 38 categories. We call this larger dataset as the CO-SKELARGE dataset. Efficient object skeletonization in such large datasets has been made possible in this paper through our speeded-up extension proposed in this paper. In the speeded-up extension, priors of key images are computed first, and then they are propagated to other images. Such an approach speeds up the process significantly. Moreover, since our method doesn't depend on annotations exclusively, we perform extensive experiments on various datasets using all the three approaches: weakly-supervised (only category-label annotations), supervised (skeleton and segmentation annotations) and unsupervised (no annotations). The proposed method achieves promising results in all three scenarios.
    
    We would like to point out that compared with our conference version \cite{8099896}, this journal submission includes the following significant extensions. 1) We construct a larger and more challenging dataset named CO-SKELARGE dataset. 2) We employ key prior propagation idea to speed up the process while handling large datasets. 3) We now provide more detailed experimental results covering all the three possible joint processing scenarios: weakly-supervised, supervised, and unsupervised.

    \section{Related Work}
    \subsection{Skeletonization}
    The research on skeletonization can be divided into three categories. First, there are some algorithms, such as ~\cite{Saha20163, Choi2003721, Shen2011196}, which can perform skeletonization if the segmented shape of the object is given. Generally, these algorithms are sensitive to distortions of the given shape. However, this problem can be tackled through methods such as pruning \cite{Shen2013}. Second, there are also some traditional image processing methods~\cite{1315062,4060951, Lindeberg1998}, which can generate skeletons by exploiting gradient intensity maps. However, they generate skeletons even for stuff such as sky, sea, and mountains. In such a case, we usually need some object mask to suppress them. Third, there are also supervised learning based methods, which require ground-truth skeletons for some images to train a model. This class of methods includes both the traditional machine learning based methods~\cite{Tsogkas2012, Shen2016306} and the recent deep learning based methods~\cite{7410521, Shen_2016_CVPR}. The performance of the traditional machine learning based methods, however, is not so satisfactory due to the limited feature learning capability in homogeneous regions. On the contrary, the recent deep learning based methods have made remarkable progress in the skeletonization as reported in \cite{Shen_2016_CVPR,shen2017deepskeleton}.Somewhat similar to the skeletonization, there are also results reported on key-point detection in \cite{he2017mask}, which is an extension of \cite{7485869}. However, it is all at the cost of requiring a complex training process on a substantial amount of annotated data (including key-points). Note that such methods are functional only in the supervised scenario. In contrast, our method depends on joint processing and can function in all three possible scenarios: weakly-supervised, supervised, and unsupervised. Also, such methods can reliably be applied only to the object categories that have been included during the training process. In contrast, our joint processing based method can work on any category. If there is no annotated data available for a particular category, we can always take the weakly-supervised perspective and apply our method. And, if images are not categorized also, we can then take the unsupervised perspective.   
    
    \subsection{Segmentation}
    Image segmentation is a classical problem, and there are many types of approaches like interactive segmentation~\cite{rother2004grabcut,6751330}, image co-segmentation~\cite{7484309,dai2013cosegmentation,kotesicip2}, semantic segmentation~\cite{7478072}, etc. While interactive segmentation needs human efforts, image co-segmentation exploits inter-image commonness prior to help segment the individual image. Different from foreground extraction that deals only with binary labels, semantic image segmentation deals with multiple labels for giving semantic meaning to each pixel. In the past few years, deep learning based methods such as fully convolution networks (FCN) have greatly advanced the performance of semantic image segmentation. Later, to decrease the annotation burden, \cite{Lin_2016_CVPR} proposed a joint framework to combine scribble-based interactive segmentation with FCN based semantic segmentation \cite{7478072} so that they can assist each other. In a similar spirit, in this work, we propose to couple two tasks named co-skeletonization and co-segmentation for mutual assistance. Co-segmentation is another type of segmentation where we jointly segment multiple images while exploiting the commonness prior existing across them. It was introduced by Rother et al. \cite{rother2006cosegmentation} using histogram matching for accomplishing this task. Since then, many co-segmentation methods have been proposed to either improve the segmentation in terms of accuracy and processing speed \cite{batra2010icoseg,hochbaum2009efficient,joulin2010discriminative,mukherjee2009half,yuan2012discovering,zhao2010mining} or scale from image pair to multiple images \cite{kim2011distributed,joulin2012multi,rubinstein2013unsupervised,faktor2013co}. Inspired by co-segmentation, we have introduced a new task called co-skeletonization that also exploits commonness but for the purpose of skeletonization. And since co-segmentation can provide the required shape in co-skeletonization, we couple these two joint processing tasks.

    \section{Proposed Method}
    In this section, we propose our joint framework for co-skeletonization and co-segmentation. The idea is to use skeletons generated from co-skeletonization as scribbles required in co-segmentation and use segmentations generated from co-segmentation as shapes required in co-skeletonization. In this manner, the two tasks become interdependent of each other. Meanwhile, commonness across the images is exploited to generate co-skeleton and co-segment priors using structure-preserving dense correspondence. Thus, empowered by these priors and interdependence between the two tasks, we jointly optimize our co-skeletonization and co-segmentation problems while taking into consideration their respective smoothness constraints.   
    
    \begin{figure*}[t]
        \begin{center}
            \includegraphics[width=0.9\linewidth]{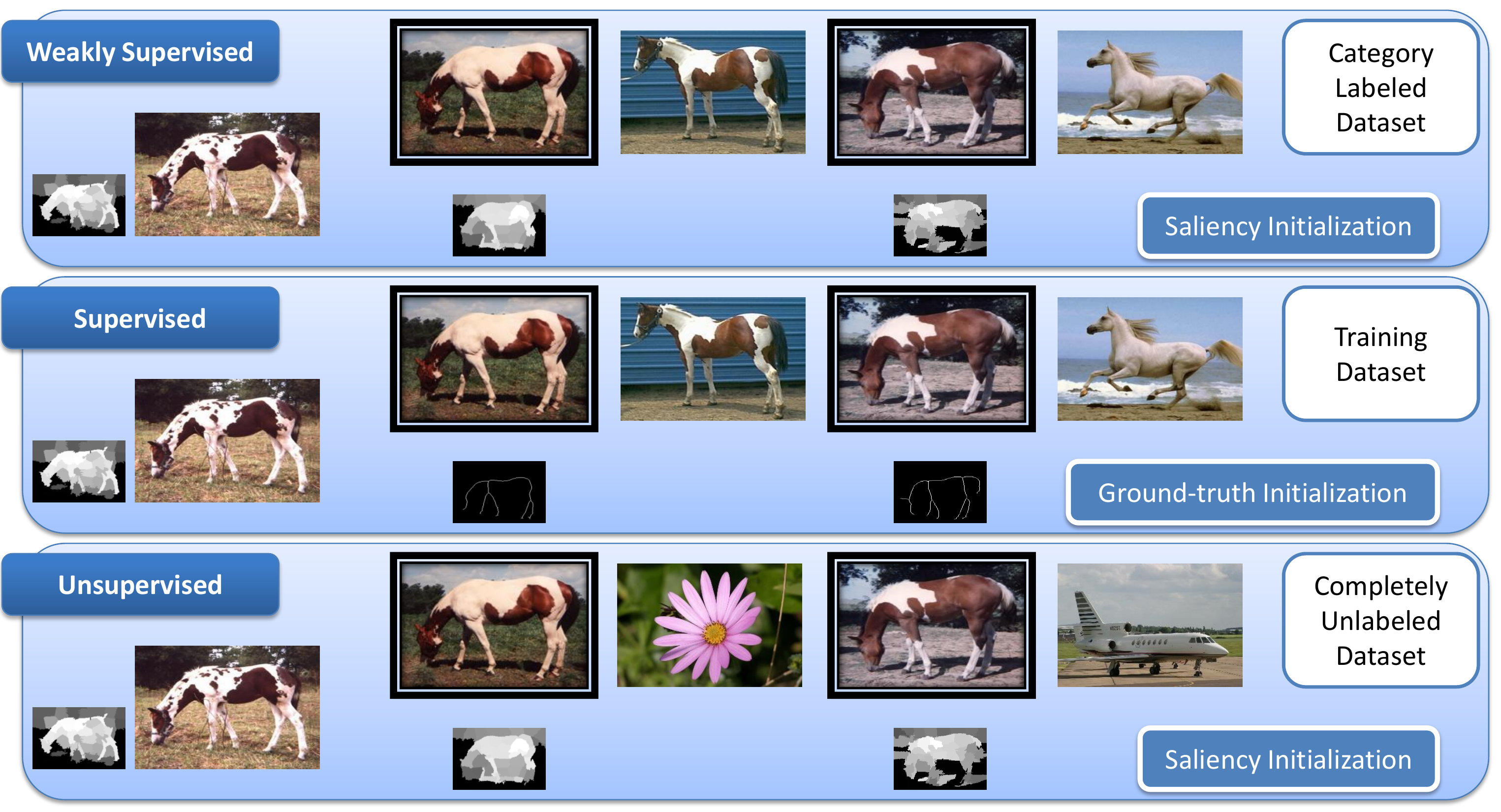}
        \end{center}
        \vspace{-1mm}
        \caption{Our method works in all three scenarios: weakly-supervised, supervised and unsupervised. The dataset composition and our initialization vary accordingly. Bordered images are the selected neighbors for the considered image.}
        \label{fig:scenario}
        
    \end{figure*}

    \subsection{Overview of Our Approach}
    Given a set of $m$ images,
    denoted by $\mathcal{I}=\{I_1,I_2,\cdots,I_m\}$, we aim to create
    two output sets: $\mathcal{K}=\{K_1,K_2,\cdots,K_m\}$ and
    $\mathcal{O}=\{O_1,O_2,\cdots,O_m\}$, comprising of skeleton masks
    and segmentation masks, respectively, where
    $K_i(p),O_i(p)\in\{0,1\}$ indicating whether a pixel $p$ is a
    skeleton pixel ($K_i(p)=1$) and whether it is a foreground pixel
    ($O_i(p)=1$).
    
    Our overall objective function for an image $I_i$ is defined as
    \begin{equation}\label{obj}
    \begin{aligned}
    \min_{K_i,O_i} &\lambda\psi_{pr}(K_i,O_i|\mathcal{N}_i)
    +\psi_{in}(K_i,O_i|I_i)+\psi_{sm}(K_i,O_i|I_i)\\ s.t. &\,\, K_i
    \subseteq \mathbf{ma}(O_i)
    \end{aligned}
    \end{equation}
    where the first term $\psi_{pr}$ accounts for the inter-image priors from a set of neighbor images denoted as $\mathcal{N}_i$, the second term
    $\psi_{in}$ is to enforce the interdependence between the skeleton
    $K_i$ and the shape/segmentation $O_i$ in image $I_i$, the third term $\psi_{sm}$ is the smoothness term to enforce smoothness, and $\lambda$ is a parameter to control
    the influence of the inter-image prior term. The constraint
    in~\eqref{obj} means the skeleton must be a subset of the medial axis
    ($\mathbf{ma}$)~\cite{Choi2003721} of the shape. Regarding neighbor images note that they are obtained using different approaches in different scenarios. In the weakly-supervised scenario, they are obtained using the k-means clustering approach on the respective category of the dataset. In the supervised scenario, they are collected using the kNN approach on the training dataset. In the unsupervised scenario, they are obtained using the k-means clustering approach on the entire dataset. 
    
    We resort to the typical alternative optimization strategy to
    solve~\eqref{obj}, i.e., dividing~\eqref{obj} into two sub-problems
    and solve them iteratively. In particular, one sub-problem is as follows. Given the shape $O_i$, we solve co-skeletonization by
    \begin{equation}\label{coskl}
    \begin{aligned}
    \min_{K_i} & \,\,\lambda\psi_{pr}^k(K_i|\mathcal{N}_i)+\psi_{in}^k(K_i|O_i)+\psi_{sm}^k(K_i)\\
    s.t. & \,\,K_i\subseteq \mathbf{ma}(O_i).
    \end{aligned}
    \end{equation}
    The other sub-problem is that given the skeleton $K_i$, we solve
    co-segmentation by
    \begin{equation}\label{coseg}
    \min_{O_i}\lambda\psi_{pr}^o(O_i|\mathcal{N}_i)+\psi_{in}^o(O_i|K_i,I_i)+\psi_{sm}^o(O_i|I_i).
    \end{equation}
    If we treat both the inter-image prior term $\psi_{pr}^k$ and the
    shape prior term $\psi_{in}^k$ as a combined prior,
    \eqref{coskl} turns out to be a skeleton pruning problem and
    can be solved using the approach similar to~\cite{Shen2013}, where
    branches in the skeleton are iteratively removed as long as it
    reduces the overall energy. Similarly, if we combine both the inter-image
    prior $\psi_{pr}^o$ and the skeleton prior $\psi_{in}^o$ as the
    data term, \eqref{coseg} becomes a standard MRF-based
    segmentation formulation, which can be solved using
    GrabCut~\cite{rother2004grabcut}. Thus, compared with the existing
    works, the key differences of our formulation lie in the designed
    inter-image prior terms as well as the interdependence terms,
    which link the co-skeletonization and co-segmentation together.
    
    In literature, several works have taken such an alternative optimization approach while combining different complicated modalities. A representative work is \cite{rubinstein2013unsupervised} where masks and correspondences were alternatively obtained. However, one good thing about such an approach is that the neighborhoods of images become better progressively if we use the results obtained at every iteration for seeking it, as reported in the same work. Given that, we can be convinced that skeletons combined with improving joint processing give better foreground seeds for segmentation, and segmentation combined with improving joint processing gives good shapes for skeletonization. This kind of iteratively solving~\eqref{coskl} and~\eqref{coseg} requires initialization first of all. For this purpose, we propose to initialize $\mathcal{O}$ by
    Otsu thresholded saliency maps \cite{8466906} and $\mathcal{K}$ by the medial
    axis mask~\cite{Choi2003721}. From experiments, we found that \cite{8466906} gave us the best results; otherwise, our method is not dependent on a particular method. Note that, in the supervised scenario, the training images are initialized using ground-truth skeletons as initial skeletons. We use the same ground-truth skeletons to generate initial segmentation masks as well using grabcut, which are denoted by $\mathbf{gc}$. We can easily generate a bounding box required in the grabcut using a skeleton. Different scenarios have been made clear in Fig.~\ref{fig:scenario}. Alg.~\ref{apprch} summarizes our
    approach, where $(\psi_{pr}+\psi_{in}+\psi_{sm})^{(t)}$ denotes
    the objective function value of \eqref{obj} at the $t^{th}$
    iteration and $\psi_{pr}=\psi_{pr}^k+\psi_{pr}^o$,
    $\psi_{in}=\psi_{in}^k+\psi_{in}^o$,
    $\psi_{sm}=\psi_{sm}^k+\psi_{sm}^o$.

    \begin{algorithm}\label{apprch}
        \KwData{An image set $\mathcal{I}$. }
        \KwResult{ Sets $\mathcal{O}$ and $\mathcal{K}$ containing segmentations and skeletons of images in $\mathcal{I}$}

        \textbf{Initialization:}
        $\forall I_i\in \mathcal{I}$,
        
        \If{$I_i \in testset$ }{
            $O_i^{(0)}=$ Otsu thresholded saliency map and $K_i^{(0)}= \mathbf{ma}(O_i^{(0)})$}\Else{$K_i^{(0)}=$ skeleton annotation and $O_i^{(0)}=\mathbf{gc}(K_i^{(0)})$}
        
        \textbf{Process:}
        $\forall I_i\in \mathcal{I}$,
        
        \Do{$(\lambda\psi_{pr}+\psi_{in}+\psi_{sm})^{(t+1)}\leq(\lambda\psi_{pr}+\psi_{in}+\psi_{sm})^{(t)}$}{

            1) Obtain $O_i^{(t+1)}$ by solving \eqref{coseg} using \cite{rother2004grabcut} with $\mathcal{O}^{(t)}$ and $K_i^{(t)}$.
            
            2) Obtain $K_i^{(t+1)}$ by solving \eqref{coskl} using  \cite{Shen2013} with $\mathcal{K}^{(t)}$ and $O_i^{(t+1)}$, $s.t.\text{ }K_i^{(t+1)}\in\mathbf{ma}(O_i^{(t+1)})$.}
        $\mathcal{O}\gets\mathcal{O}^{(t)}$ and $\mathcal{K}\gets\mathcal{K}^{(t)}$
        \caption{Our approach for solving \eqref{obj}. Note that $\mathbf{ma(\cdot)}$ and $\mathbf{gc(\cdot)}$ denote medial axis and grabcut algorithm.}
    \end{algorithm}

    \subsection{Object Co-skeletonization}
    As shown in Alg.~\ref{apprch}, the step of object
    co-skeletonization is to obtain $K^{(t+1)}$ by
    minimizing~\eqref{coskl}, given the shape $O^{(t+1)}$ and the
    previous skeleton set $\mathcal{K}^t$. Considering the constraint
    of $K_i^{(t+1)}\in\mathbf{ma}(O_i^{(t+1)})$, we only need to
    search skeleton pixels from the medial axis pixels. We build up
    our solution based on~\cite{Shen2013}, but with our carefully
    designed individual terms for~\eqref{coskl} as explained below.

\begin{figure}
        \begin{center}
            \includegraphics[width=1\linewidth]{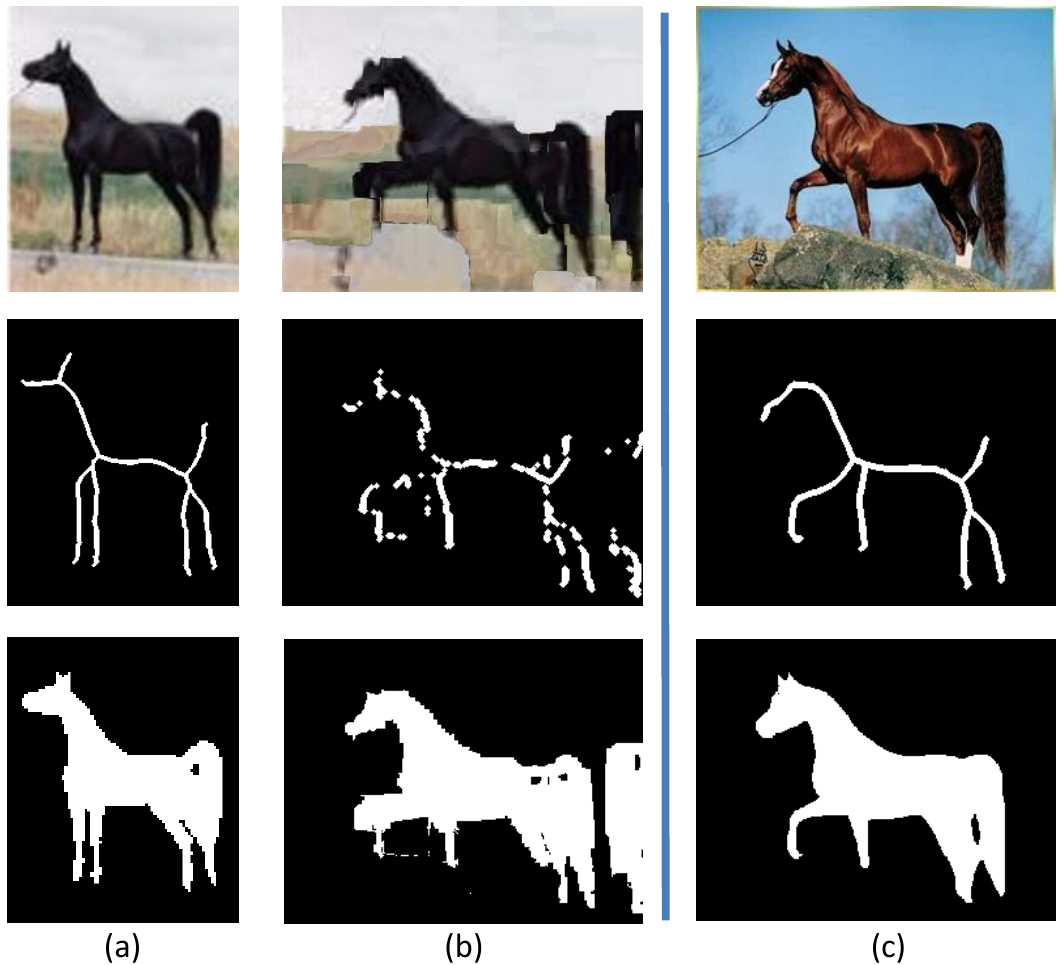}
        \end{center}
        \vspace{-1mm}
        \caption{Dense correspondences preserve the skeleton and
            segmentation structures roughly. Here (a) is warped to generate (b) to be
            used as a prior for (c).}\label{fig:coco}
    \end{figure}

    \textbf{Prior Term ($\psi^k_{pr}$):} In the object co-skeletonization, a good skeleton pixel is the one that is repetitive across images. To account for this repetitiveness, we need to find corresponding skeleton pixels in other images. However, skeleton pixels usually lie in homogeneous regions (see Fig.~\ref{fig:chal}(d)\&(e)) and are thus challenging to match. Thus, instead of trying to match sparse skeleton pixels, we make use of dense correspondences using SIFT Flow~\cite{liu2011sift}, which preserve the skeleton and segmentation structures well, as shown in Fig.~\ref{fig:coco}.

    Once the correspondence is established, we utilize the warped skeleton
    pixels from neighboring images to develop the prior term.
    Particularly, we align all the neighboring images' $t^{th}$
    iteration's skeleton maps to the concerned image $I_i$, and
    generate a co-skeleton prior at the $(t+1)^{th}$ iteration as
    \begin{equation}
    \widetilde{K}^{(t+1)}_i=\frac{K^{(t)}_i+\sum\limits_{I_j\in\mathcal{N}_i}{\mathbf{W}_j^i(K^{(t)}_j)}}{|\mathcal{N}_i|+1}\\
    \end{equation}
    where we align other skeleton maps using a warping function
    $\mathbf{W}_j^i$ \cite{liu2011sift} and then average them with
    $I_i$'s own skeleton map. Note that neighbors
    $\mathcal{N}_i$  are obtained in the GIST feature
    ~\cite{oliva2001modeling} domain. For simplicity, we drop the
    superscriptions such as $(t+1)$ in all the following derivations.
    
    Considering that the corresponding skeleton pixels from other
    images may not exactly align with the skeleton pixels of the
    considered image, we define our inter-image prior term as
    \begin{equation}\label{eq:crossentropy}
    \begin{aligned}
    \psi_{pr}^k (K_i|\mathcal{N}_i)=\frac{-\sum\limits_{p\in D_i}K_i(p)\log\Big(1+\sum\limits_{q\in
            \mathbb{N}(p)}\widetilde{K}_i(q)\Big)}{\sum\limits_{p\in D_i}K_i(p)}.
    \end{aligned}
    \end{equation}
    Eq.~\eqref{eq:crossentropy} essentially measures the consistency
    among image $I_i$'s own skeleton mask and the recommended skeleton
    mask from its neighbor images. Note that $D_i$ represents pixel domain of $I_i$ and we accumulate the
    co-skeleton prior scores in a certain neighborhood $\mathbb{N}(p)$
    for each pixel $p$ to account for the rough skeleton alignment
    across the images.

\begin{figure}
        \begin{center}
            \includegraphics[width=1\linewidth]{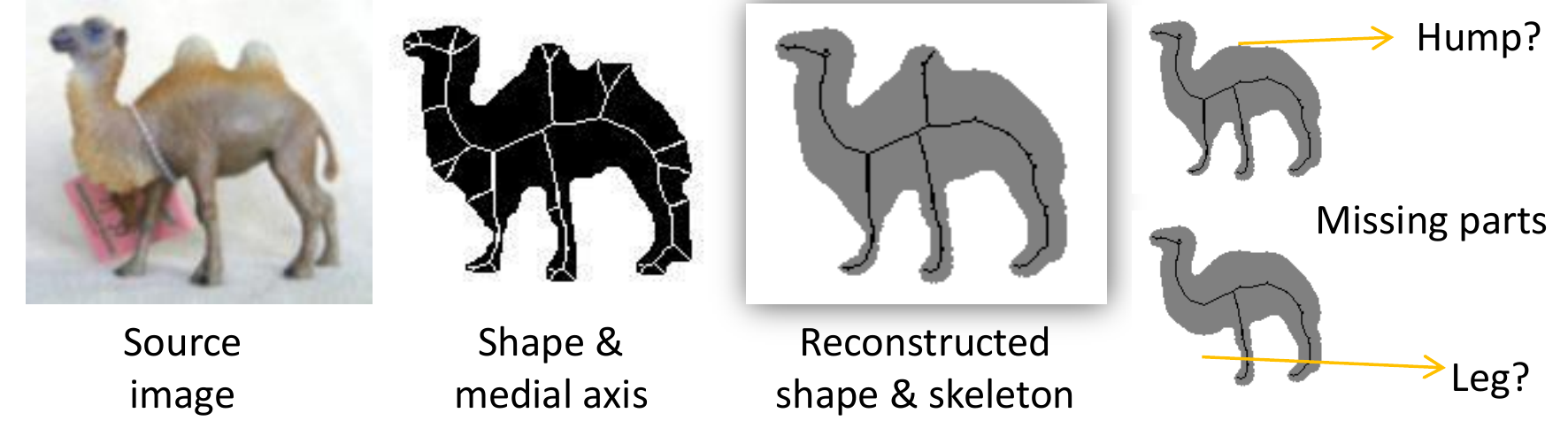}
        \end{center}
        \vspace{-1mm}
        \caption{Shape reconstruction from skeleton. Compared to the reconstructed shape from the medial axis (2nd column), the reconstructed shape (3rd column) from our simplified skeleton is simpler and smoother while still preserving the main structure. Nevertheless, we do not want an over-simplified skeleton, which will result in missing important parts in the corresponding shape reconstruction (4th column).}\label{fig:camel}
    \end{figure}

    
    \textbf{Interdependence Term ($\psi^k_{in}$):} Our interdependence term is similar to the traditional data term in skeleton pruning, i.e., it enforces that the skeleton should provide a good reconstruction of the given shape, which medial axis already does well. However, a medial axis often contains spurious branches, while the noisy shapes obtained from imperfect co-segmentation only make this worse. To avoid spurious branches, we prefer a simplified skeleton whose reconstructed shape is expected to be smooth while still
    preserving the main structure of the given shape (see
    Fig.~\ref{fig:camel} for example). On the other hand, we do not
    want an over-simplified skeleton whose reconstructed shape is likely
    to miss some important parts (see the 4th column of
    Fig.~\ref{fig:camel}).
    Therefore, we expect the reconstructed shape from the skeleton to
    match the given shape, but not necessary to be exactly the same as
    the given shape. In this spirit, we define our interdependence
    term $\psi_{in}^k$ as
    \begin{equation}
    \psi_{in}^k(K_i|O_i)=-\alpha\log{\frac{|\mathbf{R}(K_i,O_i)\cap
            O_i|}{|\mathbf{R}(K_i,O_i)\cup O_i|}}
    \end{equation}
    where we use IoU to measure the closeness between the
    reconstructed shape $\mathbf{R}(K_i,O_i)$ and the given shape
    $O_i$, and $\alpha$ is the normalization factor as defined in
    \cite{Shen2013}. Note that this term has been differently defined compared to the ~\cite{Shen2013}, which uses the simple difference between the two maps. The reconstructed shape $\mathbf{R}(K_i,O_i)$ is
    basically the union of maximal disks at skeleton
    pixels~\cite{Shen2013}, i.e.,
    \begin{equation} \label{eq:recon}
    \mathbf{R}(K_i,O_i)=\bigcup_{p\in K_i|K_i(p)=1}d(p,O_i)
    \end{equation}
    where $d(p,O_i)$ denotes the maximal disk at skeleton pixel $p$
    for the given $O_i$, and the maximal disk is the disk that exactly
    fits within $O_i$ with skeleton pixel $p$ as the center.
    
    \textbf{Smoothness Term ($\psi^k_{sm}$):} To ensure a smoother and
    simpler skeleton, we aim for a skeleton whose: (i) branches are less
    in number and (ii) branches are long. Our criteria discourage
    skeletons with spurious branches while at the same time
    encouraging skeletons with structure-defining branches. This is
    different from the criteria in~\cite{Shen2013} which only aims for
    less number of skeleton pixels. Specifically, we define the
    smoothness term $\psi_{sm}^k$ as
    \begin{equation}
    \psi_{sm}^k(K_i)=|\mathbf{b}(K_i)|\times\sum_{u=1}^{|\mathbf{b}(K_i)|}\frac{1}{length\big(b_u(K_i)\big)}
    \end{equation}
    where
    $\mathbf{b}(K_i)=\{b_1(K_i),\cdots,b_{|\mathbf{b}(K_i)|}(K_i)\}$
    denotes the set of branches of the skeleton $K_i$. In this way, we
    punish skeletons having either large number of branches (through $|\mathbf{b}(K_i)|$) or having
    short branches \Big(through $1/length\big(b_u(K_i)\big)$\Big).
    
    \subsection{Object Co-segmentation}
    The object co-segmentation problem here is as follows. Given the skeleton $K_i$, find the optimal $O_i$ that minimizes the objective function defined in Eq.~\eqref{coseg}. The individual terms in Eq.~\eqref{coseg} are defined in the following manner.

    \textbf{Prior Term ($\psi^o_{pr}$):}
    We generate an inter-image co-segment prior, similar to that for co-skeletonization. In particular, we align
    segmentation masks of neighboring images and fuse them with that
    of the concerned image, i.e.,
    \begin{equation}
    \widetilde{O}_i=\frac{O_i+\sum\limits_{I_j\in\mathcal{N}_i}{\mathbf{W}_j^i(O_j)}}{|\mathcal{N}_i|+1}\\
    \end{equation}
    where $\mathbf{W}_j^i$ is the same
    warping function from image $j$ to image $i$ as used object co-skeletonization.
    Then, with the help of $\widetilde{O}_i$, we define our inter-image prior term as
    \begin{equation}
    \begin{aligned}
    \psi_{pr}^o(O_i|\mathcal{N}_i) & =\sum_{p\in D_i}-\Bigg(O_i(p)\log\Big(\frac{1}{|\mathbb{N}(p)|}\sum\limits_{q\in\mathbb{N}(p)}\widetilde{O}_i(q)\Big)\\
    &+\Big(1-O_i(p)\Big)\log\Big(1-\frac{1}{|\mathbb{N}(p)|}\sum\limits_{q\in\mathbb{N}(p)}\widetilde{O}_i(q)\Big)\Bigg)
    \end{aligned}
    \end{equation}
    which encourages the shape to be consistent with $\widetilde{O}_i$. Here again we account for pixel correspondence errors by
    neighborhood $\mathbb{N}(p)$ averaging in the pixel domain $D_i$. 
    
    \textbf{Interdependence Term ($\psi^o_{in}$):}
    For the co-segmentation process to benefit from
    co-skeletonization, our basic idea is to build
    foreground and background appearance models based on the given skeleton $K_i$.
    Notably, we use GMM for appearance models. The foreground GMM model is learned using $K_i$ (i.e., treating skeleton pixels as foreground seeds), whereas the background GMM is learned using the background part of $ K_i$'s reconstructed shape $\mathbf{R}(K_i, O_i)$. In this manner, the appearance model is developed entirely using the skeleton. Note that, in the beginning, it is not robust to build up the GMM appearance models in this manner since the initial skeleton extracted based on saliency is not reliable at all. Thus, at initialization, we develop the foreground and background appearance models based on the inter-image priors $\widetilde{K}_i$ and $\widetilde{O}_i$, respectively.
    
    Denoting $\theta(K_i,I_i)$ as the developed appearance models, we define the interdependence term $\psi_{in}^o$ as
    \begin{equation}
    \psi_{in}^o(O_i |K_i,I_i)=\sum_{p\in
        D_i}-\log\bigg(P\Big(O_i(p)\,\mathlarger{|}\,\theta(K_i,I_i),I_i(p)\Big)\bigg)
    \end{equation}
    where $P\Big(O_i(p)\,\mathlarger{|}\,\theta(K_i,I_i),I_i(p)\Big)$
    denotes how likely a pixel of color $I(p)$ will take the label
    $O_i(p)$ given $\theta(K_i,I_i)$.
    $\psi_{in}^o$ is similar to the data term in the
    interactive segmentation method~\cite{rother2004grabcut}.
    
    \textbf{Smoothness Term ($\psi^o_{sm}$):} For ensuring smooth foreground and background segments, we simply
    adopt the smoothness term of GrabCut~\cite{rother2004grabcut}, i.e.,
    \begin{equation} \label{eq:phi_sm}
    \psi_{sm}^o(O_i|I_i)=\gamma\sum_{(p,q)\in E_i}{[O_i(p)\neq
        O_i(q)]e^{(-\beta||I_i(p)-I_i(q)||^2)}}
    \end{equation}
    where $E_i$ denotes the set of neighboring pixel pairs in the image
    $I_i$, and $\gamma$ and $\beta$ are segmentation smoothness related
    parameters as discussed in~\cite{rother2004grabcut}.


    \subsection{Key Prior Propagation}
    
    In the approach discussed so far, particularly in weakly-supervised and unsupervised scenarios, neighbors were other members in a cluster. Therefore, the interaction effectively involved aligning each image \emph{w.r.t.} each image in a cluster for generating the individual priors. Note that these alignments are computationally expensive. So, the current approach needs some modification while dealing with large datasets. To overcome this problem, we assume that the alignment is precise, which means that the same sets of corresponding pixels get together
    every time we try to collect them for different images in
    the cluster. In case this assumption holds, collecting the pixels for each image appears to be repetitive. We can speed up this process by collecting them only once and propagate back the generated priors to others. We call such priors as key priors. For this purpose, for any cluster, we choose the nearest image of the cluster-center as the key image, say $I_n$ in the $n^{th}$ cluster $\mathcal{C}_n$, for which we first get corresponding pixels and compute the key priors. Then we propagate these priors to other cluster members by aligning back for forming their priors. 
    Specifically, we get key co-skeleton prior at any iteration using \begin{equation}
    \widetilde{K}^{(t+1)}_n=\frac{K^{(t)}_n+\sum\limits_{I_j\in\mathcal{C}_n, I_j\neq I_n}{\mathbf{W}_j^n(K^{(t)}_j)}}{|\mathcal{C}_n|}\\
    \end{equation}, and the key co-segment prior is obtained as \begin{equation}
    \widetilde{O}^{(t+1)}_n=\frac{O^{(t)}_n+\sum\limits_{I_j\in\mathcal{C}_n, I_j\neq I_n}{\mathbf{W}_j^n(O^{(t)}_j)}}{|\mathcal{C}_n|}.\\
    \end{equation}
    
    As far as co-skeleton and co-segment priors of other members $\{I_j|I_j\in \mathcal{C}_n, I_j\neq I_n\}$ are concerned, they are computed using 
    \begin{equation}
    \widetilde{K}^{(t+1)}_j=\mathbf{W}_n^j(\widetilde{K}^{(t)}_n) \text{ and } \widetilde{O}^{(t+1)}_j=\mathbf{W}_n^j(\widetilde{O}^{(t)}_n),\\
    \end{equation}
    respectively. In our original approach, in a cluster $C_n$, every time we collected corresponding pixels for an
    image, we required $(|\mathcal{C}_n| - 1)$ alignments for one image in the cluster. Thus, total number of alignments required in a cluster becomes $|\mathcal{C}_n|(|\mathcal{C}_n| - 1)$. But after this modification, we need to align $(\mathcal{C}_n - 1)$ times only for the key image, and then we need another $(|\mathcal{C}_n|-1)$ alignments for propagation. Thus, the total number of alignments now turns out to be only
    $2(|\mathcal{C}_n|-1)$. There is a clear speed up of $|\mathcal{C}_n|/2$ from $|\mathcal{C}_n|(|\mathcal{C}_n| - 1)$ to $2(|\mathcal{C}_n|-1)$ in the number of alignments required using our key prior propagation approach. Also, we speed-up our approach from having a quadratic time-complexity to one having only a linear time complexity.

    \subsection{Implementation Details}
    We use the saliency extraction method \cite{6909756} for
    initialization of our framework in our experiments. Note that the proposed method is not restricted to this particular saliency extraction method; any other method can also be used for the initialization purpose. We use the same default setting as reported in \cite{rother2004grabcut} for the segmentation parameters
    $\gamma$ and $\beta$ in~\eqref{eq:phi_sm} throughout our experiments. For the parameters of SIFT flow~\cite{liu2011sift}, we follow the setting in~\cite{rubinstein2013unsupervised} to handle the possible matching of different semantic objects. The parameter $\lambda$ in both~\eqref{coskl} and~\eqref{coseg}, which controls the influence of joint processing, is set to 0.1.
    
    \section{Experimental Results}
    
    \begin{figure*}
        \begin{center}
            \includegraphics[width=1\linewidth]{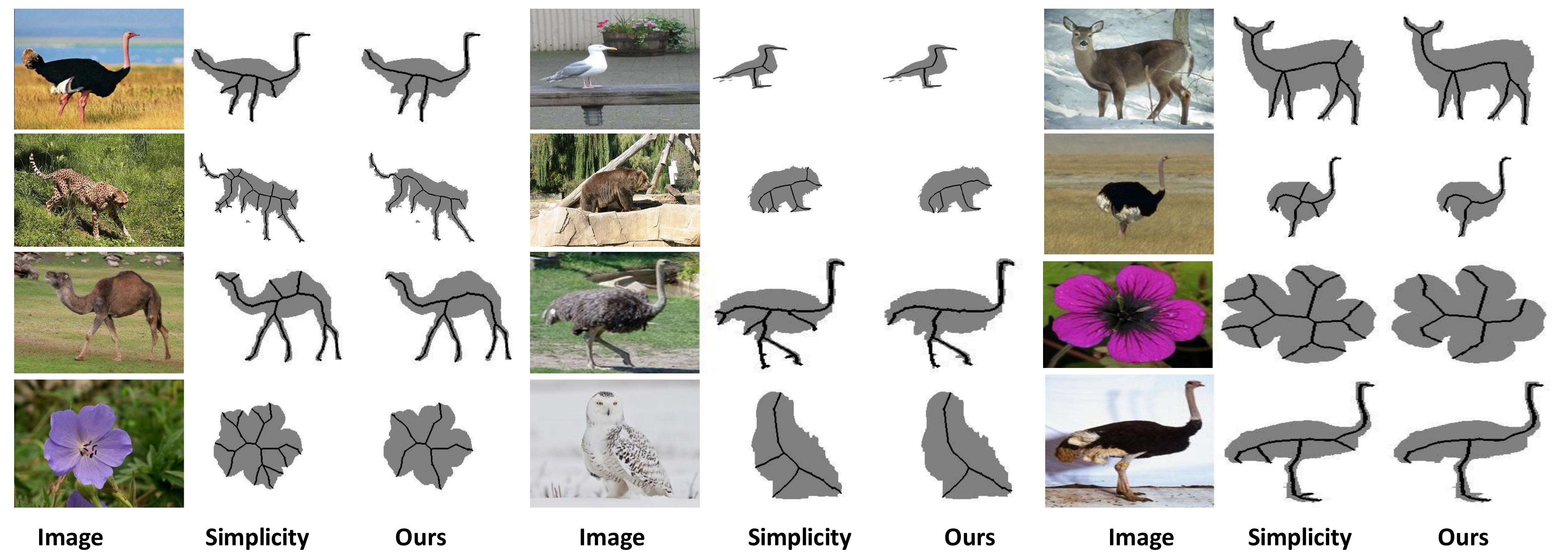}
        \end{center}
        \vspace{-2mm}
        \caption{Given the shape, we improve skeletonization method Smplicity\cite{Shen2013} using our improved terms in Simplicity\cite{Shen2013}'s objective function. It can be seen that our skeletons are much smoother and better in representing the shape.}\label{fig:dataset}
    \end{figure*}
    
    In this section, we first discuss the different datasets and evaluation metrics used in our experiments. Then, we discuss our experiments in three joint processing scenarios, namely weakly-supervised, supervised, and unsupervised. Lastly, we provide some discussions and limitations. We generally report our results using the original approach unless specified, i.e., we specify whenever we resort to speed-up through key-prior generation.     
    
    \subsection{Datasets and Evaluation Metrics}
    \textbf{Datasets:} So far, there is only one publicly available dataset, i.e.
    WH-SYMMAX dataset~\cite{Shen2016306}, on which weakly supervised co-skeletonization can be performed, but it contains only the horse category of images. To evaluate the weak-supervised co-skeletonization task extensively, we develop a new benchmark dataset named CO-SKELARGE dataset. Its initial version named CO-SKEL was presented in our preliminary work \cite{8099896}. While the CO-SKEL dataset consisted of 26 categories with a total of 348 images of animals, birds, flowers, and humans, the CO-SKELARGE dataset consists of 38 categories and a total of 1831 images. These images are collected from various publicly available datasets such as MSRC, CosegRep, and Weizmann Horses along with their ground-truth segmentation masks. We apply~\cite{Shen2013}
    (with our improved terms) on these ground-truth masks, just like the way WH-SYMMAX dataset was developed from the Weizmann Horses dataset~\cite{Borenstein2002}. Fig.~\ref{fig:dataset} shows some example images, and their skeletons using~\cite{Shen2013} and our improvement
    of~\cite{Shen2013}. Thanks to our improved terms, it can be seen that our skeletons are much smoother and better in representing the shapes.

    \textbf{Metrics:} For evaluation of co-skeletonization, we calculate
    F-measure (which combines precision and recall). 
    Considering it is challenging to get a resultant skeleton mask exactly aligned with the ground-truth, it can be considered as a hit if a resultant skeleton pixel is within a tolerable distance from the ground-truth skeleton pixel. Therefore, we consider a resultant skeleton pixel as correct if it is within $d\in\{0,1,2,3,4,5\}$ pixels from a ground-truth skeleton pixel, and report $F^d$ as the corresponding F-measure score. We do this to cover a range of tolerance levels to see how the performance varies. Previous works reported their results with the $d$ adaptive to size of the image, i.e, $d=0.0075\times\sqrt{width^2+height^2}$. We denote this metric as $F^\alpha$.

    \subsection{Weakly Supervised Co-skeletonization Results}
    
    \begin{table}
        \renewcommand{\tabcolsep}{5pt}
        \begin{center}
            \begin{tabular}{|l|cccccc|}
                \hline
                & $F^0$ & $F^1$ & $F^2$ & $F^3$ & $F^4$& $F^5$ \\
                \hline\hline
                Ours$^{(0)}$ &0.098& 0.234&0.294&0.345&0.407&0.448\\
                Ours (w/o $in$) &0.178&0.362& 0.419 &0.462&0.509&0.538\\\hline
                Ours & \textbf{0.216} &\textbf{0.435}&\textbf{0.502}&\textbf{0.554}&\textbf{0.610}&\textbf{0.643}\\
                \hline
            \end{tabular}
        \end{center}
        \caption{Comparisons of weakly-supervised co-skeletonization results of our method and its two baselines on the WH-SYMMAX dataset. Ours$^{(0)}$: our initialization baseline using Otsu thresholded saliency maps~\cite{cheng2011global} and medial axis approach. Ours (w/o $in$): our method without using the interdependence, i.e. running co-segmentation followed by skeletonization.}\label{tab:horses}
    \end{table}
    
    \begin{table}
        \renewcommand{\tabcolsep}{5pt}
        \begin{center}
            \begin{tabular}{|l|cccccc|}
                \hline
                & $F^0$ & $F^1$ & $F^2$ & $F^3$ & $F^4$& $F^5$ \\
                \hline\hline
                Ours$^{(0)}$ &0.141 &0.299&0.361&0.406&0.458&0.490\\
                Ours (w/o $in$) & 0.212 &0.448&0.518&0.561&0.605&0.630\\\hline
                Ours & \textbf{0.237}    &\textbf{0.457}    &\textbf{0.524}    &\textbf{0.571} &  \textbf{0.619} &   \textbf{0.647}\\
                \hline
            \end{tabular}
        \end{center}
        \caption{Comparisons of the weakly-supervised co-skeletonization results of our method and its two baselines on our CO-SKEL dataset.}\label{tab:coskel}
    \end{table}
    \begin{table}
        \renewcommand{\tabcolsep}{5pt}
        \begin{center}
            \begin{tabular}{|l|cccccc|}
                \hline
                & $F^0$ & $F^1$ & $F^2$ & $F^3$ & $F^4$& $F^5$ \\
                \hline\hline
                Ours$^{(0)}$ &0.118 &0.274&0.339&0.388&0.443&0.477\\
                Ours (w/o $in$) & 0.160 &0.362&0.430&0.475&0.522&0.549\\\hline
                Ours & \textbf{0.178}    &\textbf{0.374} &   \textbf{0.439}    &\textbf{0.486} &   \textbf{0.534} &   \textbf{0.562}\\
                \hline
            \end{tabular}
        \end{center}
        \caption{Comparisons of the weakly-supervised co-skeletonization results of our method and the two baselines on our CO-SKELARGE dataset.}\label{tab:coskelarge}
    \end{table}

    \begin{figure*}
        \begin{center}
            \includegraphics[width=1\linewidth]{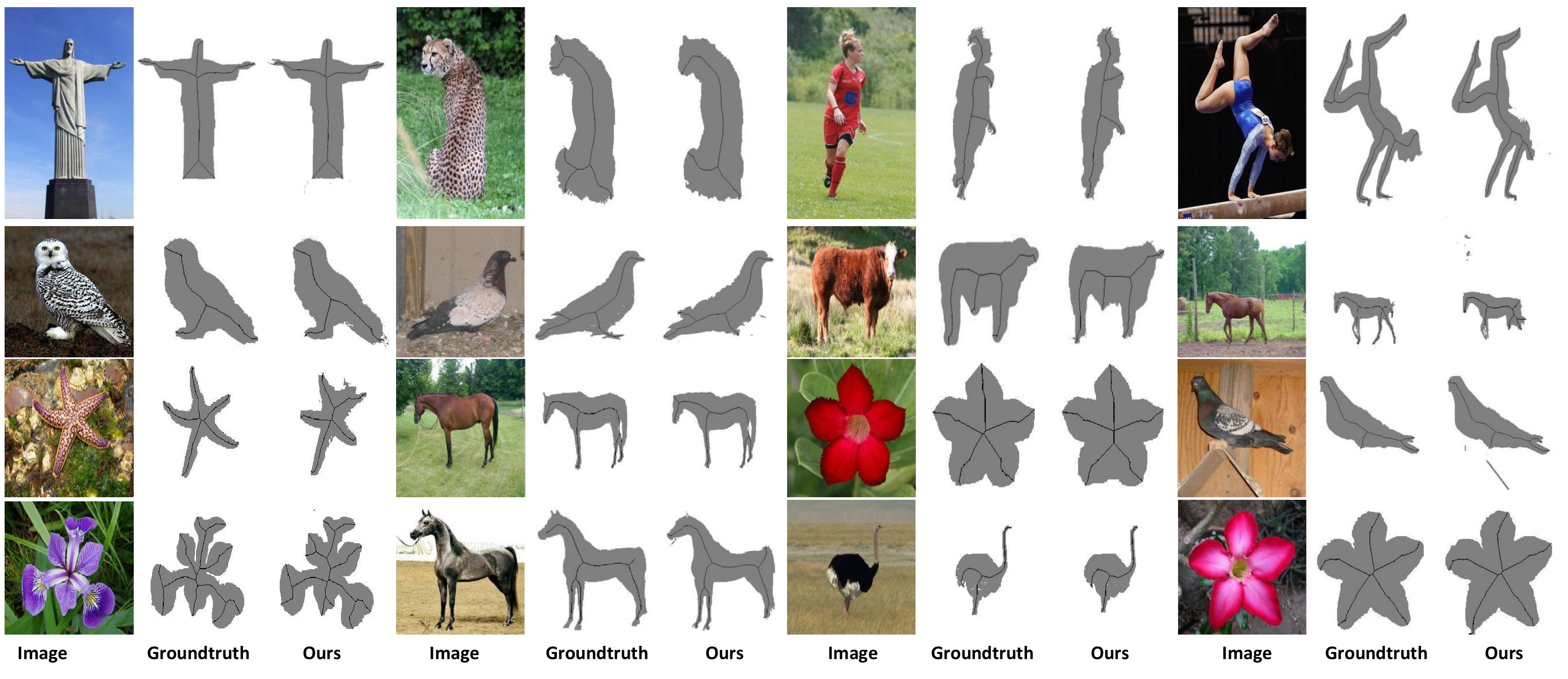}
        \end{center}
        \caption{Sample weakly-supervised co-skeletonization results on CO-SKEL dataset along with our final shape masks. It can be seen that both are quite close to the groundtruths.}\label{fig:visres}
    \end{figure*}
    
    \begin{figure*}
        \begin{center}
            \includegraphics[width=1\linewidth]{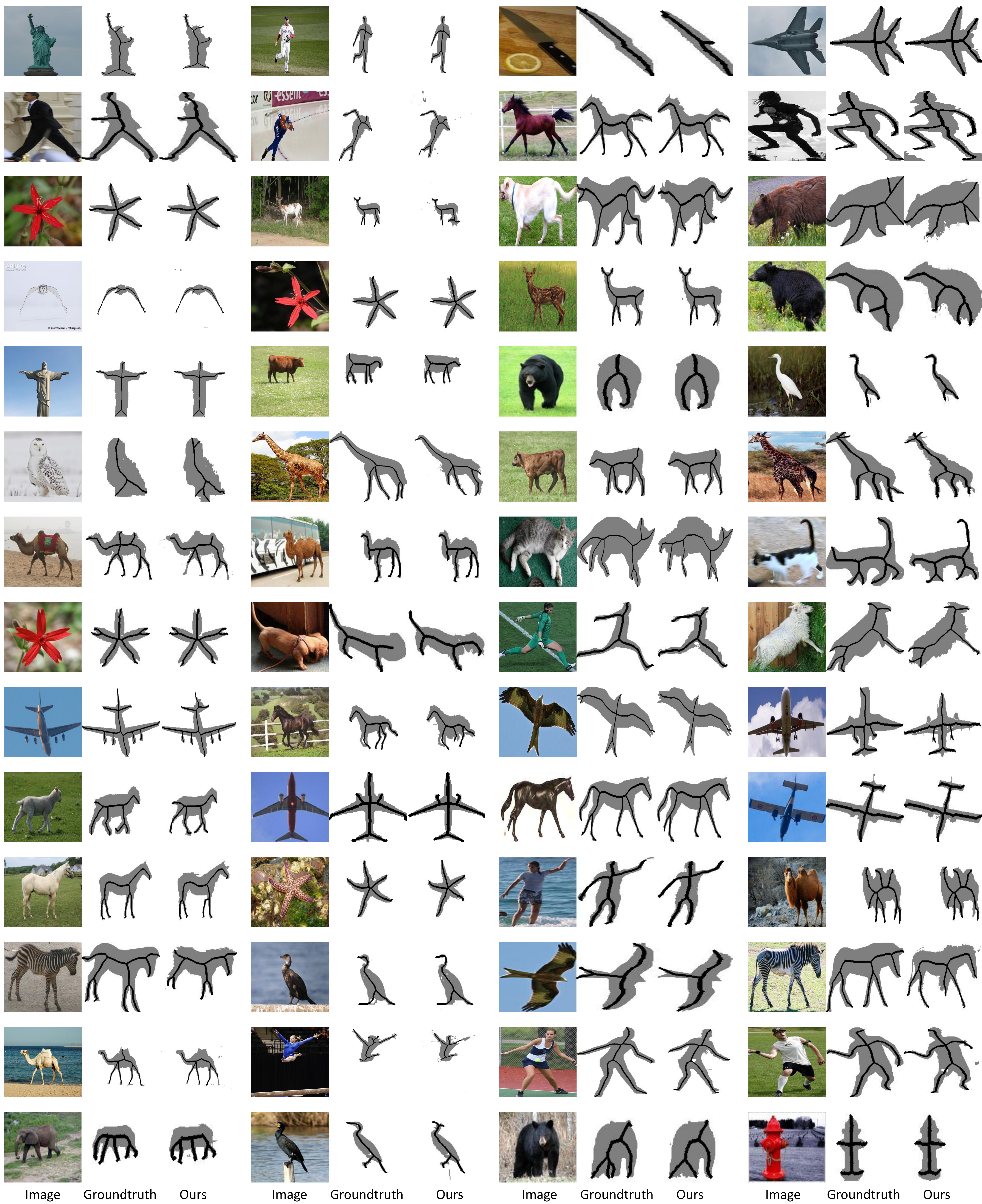}
        \end{center}
        \caption{Sample weakly-supervised co-skeletonization results on CO-SKELARGE dataset along with our final shape masks. It can be seen that both are quite close to the groundtruths.}\label{fig:coskelarge}
    \end{figure*}
    
    \begin{table}[]
        \renewcommand{\tabcolsep}{5pt}
        \begin{center}
            \begin{tabular}{|l|c|cccccc|}
                \hline
                &$m$& $F^0$ & $F^1$ & $F^2$ & $F^3$ & $F^4$ &$F^5$ \\
                \hline\hline
                bear    &4& 0.050 & 0.189 & 0.259 & 0.305 & 0.379 & 0.425 \\
                camel    &10& 0.377 & 0.618 & 0.686 & 0.727 & 0.761 & 0.781 \\
                cat    &8& 0.143 & 0.419 & 0.530 & 0.610 & 0.679 & 0.724 \\
                cheetah    &10& 0.069 & 0.179 & 0.239 & 0.297 & 0.361 & 0.397 \\
                cormorant    &8& 0.346 & 0.543 & 0.599 & 0.647 & 0.698 & 0.730 \\
                cow    &28& 0.138 & 0.403 & 0.530 & 0.615 & 0.691 & 0.733 \\
                cranesbill    &7& 0.240 & 0.576 & 0.658 & 0.706 & 0.748 & 0.774 \\
                deer    &6& 0.279 & 0.465 & 0.522 & 0.577 & 0.649 & 0.683 \\
                desertrose    &15& 0.329 & 0.675 & 0.755 & 0.800 & 0.838 & 0.857 \\
                dog    &11& 0.133 & 0.410 & 0.504 & 0.573 & 0.637 & 0.670 \\
                erget &14    & 0.405 & 0.626 & 0.665 & 0.696 & 0.723 & 0.742 \\
                firepink    &6& 0.493 & 0.839 & 0.895 & 0.925 & 0.949 & 0.958 \\
                frog    &7& 0.183 & 0.366 & 0.433 & 0.492 & 0.544 & 0.576 \\
                germanium    &17& 0.295 & 0.614 & 0.697 & 0.750 & 0.801 & 0.828 \\
                horse    &31& 0.273 & 0.491 & 0.551 & 0.599 & 0.651 & 0.683 \\
                iris    &10& 0.358 & 0.637 & 0.703 & 0.746 & 0.790 & 0.808 \\
                man    &20& 0.126 & 0.260 & 0.299 & 0.329 & 0.365 & 0.387 \\
                ostrich    &11& 0.291 & 0.514 & 0.582 & 0.625 & 0.658 & 0.679 \\
                panda    &15& 0.031 & 0.094 & 0.135 & 0.169 & 0.216 & 0.245 \\
                pigeon    &16& 0.144 & 0.288 & 0.333 & 0.367 & 0.405 & 0.425 \\
                seagull    &13& 0.180 & 0.336 & 0.397 & 0.448 & 0.500 & 0.530 \\
                seastar    &9& 0.448 & 0.739 & 0.781 & 0.811 & 0.839 & 0.851 \\
                sheep    &10& 0.115 & 0.342 & 0.465 & 0.535 & 0.608 & 0.650 \\
                snowowl    &10& 0.130 & 0.281 & 0.325 & 0.362 & 0.421 & 0.457 \\
                statue    &29& 0.301 & 0.525 & 0.571 & 0.599 & 0.623 & 0.642 \\
                woman    &23& 0.276 & 0.451 & 0.500 & 0.536 & 0.572 & 0.596 \\
                \hline
                variance&    &0.016  &  0.034 &   0.034  &  0.034 &   0.032&    0.030\\
                \hline 
            \end{tabular}
        \end{center}
        \caption{Categorywise number of images and our weakly-supervised co-skeletonization results on the CO-SKEL dataset.}\label{tab:category}
    \end{table}

    \begin{table}[]
        \renewcommand{\tabcolsep}{5pt}
        \begin{center}
            \begin{tabular}{|l|c|cccccc|}
                \hline
                &$m$& $F^0$ & $F^1$ & $F^2$ & $F^3$ & $F^4$ &$F^5$\\
                \hline\hline
                banana    &  7& 0.122 & 0.358 & 0.422 & 0.469 & 0.528 & 0.564 \\
                bear    &  24& 0.134 & 0.332 & 0.413 & 0.479 & 0.541 & 0.572 \\
                brush    &  6& 0.102 & 0.317 & 0.419 & 0.496 & 0.551 & 0.574 \\
                camel    &  12& 0.300 & 0.501 & 0.555 & 0.592 & 0.634 & 0.662 \\
                cat    &  42& 0.118 & 0.333 & 0.421 & 0.483 & 0.546 & 0.582 \\
                cheetah    & 10 & 0.067 & 0.200 & 0.265 & 0.322 & 0.387 & 0.425 \\
                cormorant    & 8 & 0.370 & 0.577 & 0.612 & 0.633 & 0.660 & 0.684 \\
                cow    & 87 & 0.121 & 0.344 & 0.445 & 0.509 & 0.572 & 0.609 \\
                cranesbill    & 7 & 0.257 & 0.563 & 0.634 & 0.679 & 0.717 & 0.744 \\
                deer    & 6 & 0.211 & 0.355 & 0.398 & 0.441 & 0.486 & 0.512 \\
                desertrose    & 15 & 0.311 & 0.630 & 0.700 & 0.740 & 0.780 & 0.800 \\
                dog    & 70 & 0.125 & 0.352 & 0.439 & 0.497 & 0.557 & 0.591 \\
                eagle    & 9 & 0.168 & 0.489 & 0.582 & 0.648 & 0.722 & 0.757 \\
                egret    &  14& 0.420 & 0.633 & 0.666 & 0.694 & 0.726 & 0.743 \\
                elephant    &46  & 0.078 & 0.214 & 0.275 & 0.321 & 0.376 & 0.410 \\
                firepink    &  6& 0.422 & 0.721 & 0.784 & 0.833 & 0.888 & 0.911 \\
                flowerwise    &  8& 0.072 & 0.209 & 0.288 & 0.347 & 0.405 & 0.442 \\
                frog    &  7& 0.180 & 0.371 & 0.425 & 0.473 & 0.513 & 0.534 \\
                geranium    & 17 & 0.285 & 0.600 & 0.691 & 0.743 & 0.791 & 0.819 \\
                giraffe    &  213& 0.100 & 0.277 & 0.346 & 0.392 & 0.442 & 0.470 \\
                horse    &  245& 0.143 & 0.348 & 0.419 & 0.467 & 0.520 & 0.552 \\
                hydrant    &  62& 0.151 & 0.360 & 0.459 & 0.520 & 0.572 & 0.605 \\
                iris    & 10 & 0.351 & 0.634 & 0.702 & 0.743 & 0.787 & 0.805 \\
                cutlery    &  6& 0.055 & 0.149 & 0.207 & 0.257 & 0.303 & 0.331 \\
                man    & 411 & 0.120 & 0.322 & 0.401 & 0.455 & 0.509 & 0.538 \\
                ostrich    & 11 & 0.304 & 0.542 & 0.608 & 0.651 & 0.688 & 0.712 \\
                panda    &  15& 0.041 & 0.119 & 0.170 & 0.215 & 0.265 & 0.297 \\
                parrot    &  5& 0.040 & 0.143 & 0.214 & 0.282 & 0.348 & 0.378 \\
                pigeon    &  19& 0.124 & 0.255 & 0.299 & 0.331 & 0.361 & 0.382 \\
                plane    &  169& 0.202 & 0.496 & 0.593 & 0.648 & 0.702 & 0.732 \\
                seagull    &  26& 0.185 & 0.364 & 0.434 & 0.483 & 0.525 & 0.553 \\
                seastar    &  9& 0.407 & 0.654 & 0.686 & 0.708 & 0.735 & 0.749 \\
                sheep    &  50& 0.093 & 0.279 & 0.359 & 0.414 & 0.474 & 0.507 \\
                snowowl    &  10& 0.110 & 0.261 & 0.315 & 0.354 & 0.402 & 0.435 \\
                statue    &  29& 0.327 & 0.576 & 0.622 & 0.646 & 0.668 & 0.684 \\
                swan    &  6& 0.086 & 0.175 & 0.206 & 0.238 & 0.277 & 0.305 \\
                woman    &  122& 0.165 & 0.376 & 0.456 & 0.511 & 0.565 & 0.595 \\
                zebra    &  12& 0.131 & 0.344 & 0.430 & 0.497 & 0.563 & 0.597 \\
                \hline
                variance    &  &0.013        &0.027        &  0.026      &   0.025     &    0.024    &  0.024     \\
                \hline
            \end{tabular}
        \end{center}
        \caption{Categorywise number of images and our weakly supervised co-skeletonization results on the CO-SKELARGE dataset.}\label{tab:category2}
    \end{table}

    \begin{figure*}
        \begin{center}
            \includegraphics[width=1\linewidth]{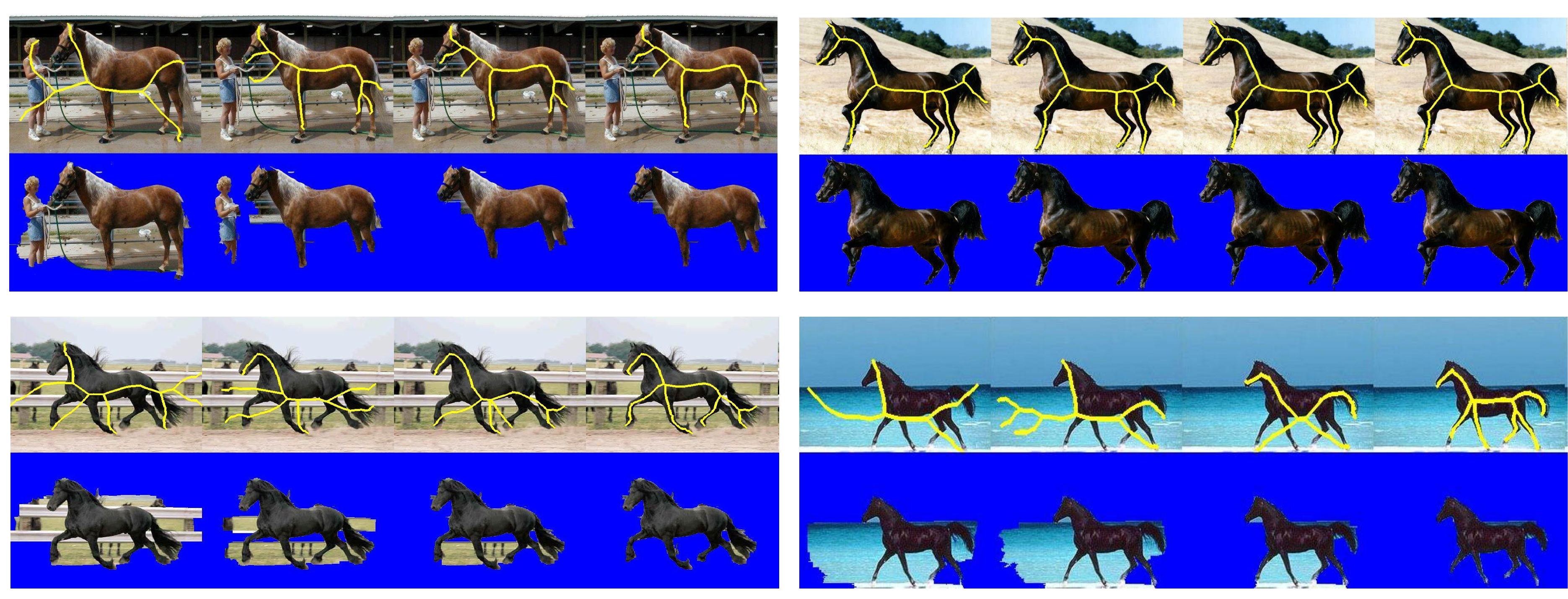}
        \end{center}
        \vspace{-1mm}
        \caption{Some examples of steadily improving skeletonization and segmentation after each iteration. The top-right example shows that our model continues to reproduce similar results once the optimal shape and skeleton are obtained.}\label{fig:horses}
    \end{figure*}
    
    \begin{figure}
        \begin{center}
            
            \includegraphics[width=1\linewidth]{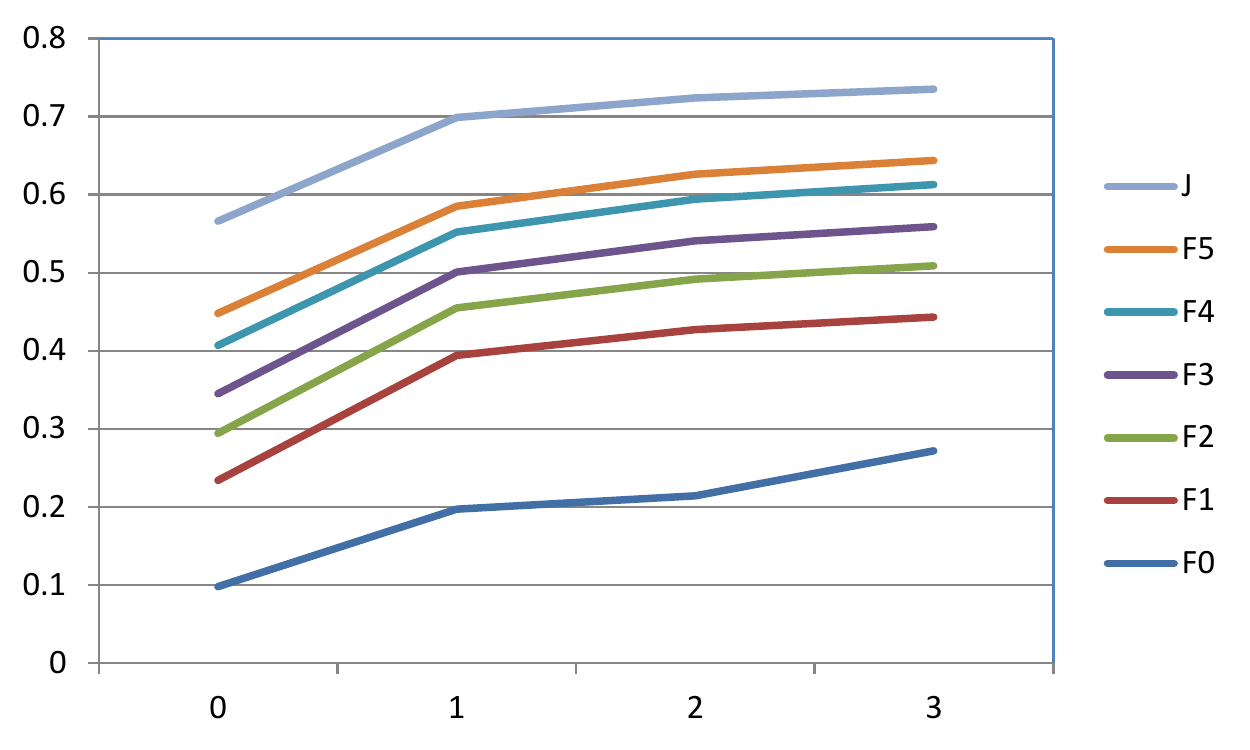}
        \end{center}
        \vspace{-1mm}
        \caption{Performance v/s Iteration plot. It can be seen that both the skeletonization performance (denoted by F0-F5) and segmentation performance (denoted by J) improve after every iteration synergistically.}\label{fig:iter}
    \end{figure}

    In the weakly-supervised scenario, with the availability of category labels, the neighbor image always has the same semantic object as in the considered image. We report our overall weakly-supervised co-skeletonization performance on WH-SYMMAX and our two datasets in Tables~\ref{tab:horses}-\ref{tab:coskelarge}. Note that the results reported here are on the entire WH-SYMMAX dataset, not just the test set. For the CO-SKELARGE dataset, considering its large size, we apply our speeded-up approach to obtain the results. For comparison, since we are the first to do skeletonization in a weakly-supervised scenario, we compare it with two baselines. One is our initialization baseline Ours$^{(0)}$, and another is Ours (w/o $in$), where we neglect the interdependence, i.e., running co-segmentation first and then performing skeletonization from the resultant foreground segments. Essentially, one baseline uses simple otsu and medial axis methods, and another uses sophisticated co-segmentation and skeletonization methods. Note that we use a saliency map \cite{6909756} for initialization and otsu mask of the co-skeleton prior itself in the interdependence term (note no use of skeleton here). It can be seen that our method outperforms both baselines on all three datasets and for all $d$. Fig.~\ref{fig:visres}-~\ref{fig:coskelarge} shows some sample results of our method along with corresponding ground-truths on our datasets. Also, we report results on individual categories of our two datasets in Tables~\ref{tab:category}-\ref{tab:category2}. Low variances suggest that our method is sufficiently reliable.

    In Fig.~\ref{fig:horses}, few samples are shown where the results improve iteration by iteration. To analyze this quantitatively, we evaluate the performance after every iteration in Fig.~\ref{fig:iter} on the WH-SYMMAX dataset. It can be seen that the performances, denoted by $F^0-F^5$ and $J$ (Jaccard Similarity for segmentation evaluation), improve after every iteration steadily and synergistically.  A choice of 2 to 3 iterations is good enough for our method to obtain reasonable performance.

    \begin{table}
        \begin{center}
            \begin{tabular}{|l|c|c|c|}
                \hline
                Methods&  WH-SYMMAX & SK506 & SK-LARGE\\
                \hline\hline
                Symmetric \cite{5459472}&0.174&0.218&0.243\\
                Deformable Disc \cite{6751328}&0.223&0.252&0.255\\
                Particle Filter \cite{6940228}&0.334&0.226&-\\
                Distance Regression \cite{6909741}&0.103&-&-\\
                MIL \cite{Tsogkas2012}&0.365&0.392&0.293\\
                MISL \cite{4060951}&0.402&-&-\\
                \hline
                Ours (S) &\textbf{0.618}&\textbf{0.525}&\textbf{0.501}\\
                \hline
            \end{tabular}
        \end{center}
        \vspace{1mm}
        \caption{Comparisons of the supervised co-skeletonization results of our method with other supervised methods using $F^\alpha$ metric.}\label{tab:supcomp}
    \end{table}

    \subsection{Supervised Co-skeletonization Results}
    In the literature, there are numerous supervised skeletonization methods available for comparison. As mentioned earlier, while using the proposed method in the supervised scenario, the training images are initialized with ground truths, and kNN is applied to find the nearest neighbors of testing images from the training dataset. Such neighbors with ground-truths naturally boost the performance of co-skeletonization for developing useful priors of the test images. We make the comparison with existing supervised skeletonization methods on test images of the WH-SYMMAX and SK506 datasets in Table~\ref{tab:supcomp}. We denote the co-skeletonization results of our supervised approach as "Ours (S)", where we provide ground-truth skeleton annotations to the training images. Note that their segment masks can be easily computed from skeletons using grabcut through the enclosing bounding box. Our supervised approach outperforms the other supervised methods convincingly. Note that we take the performance values of other methods from~\cite{Shen_2016_CVPR}. We would like to point out that the recently developed deep learning based supervised methods~\cite{Shen_2016_CVPR,shen2017deepskeleton} report better performance. However, we refrain from comparing with it for the following reasons: (i) The results that ~\cite{Shen_2016_CVPR} reports are max-pooled results obtained from tuning over a wide range of thresholds. In contrast, our results reported here are without any parameter tuning. (ii) ~\cite{Shen_2016_CVPR} takes extra help in the form of a pre-trained network, which they use to build upon, whereas we rely on just saliency estimation. (iii) ~\cite{Shen_2016_CVPR} uses both skeleton annotation and shape information annotation while training, whereas we use only skeleton annotations for training images. (iv) Essentially, our proposed method is a weakly-supervised method or unsupervised method, with the possibility of adding strong supervision, however, by replacing the saliency initialization with ground-truth skeleton initialization in the training dataset while extracting the neighbors. There is no training involved as such in such this approach, meaning we are not learning any parameters. In contrast, the existing methods are supervised ones. To compare with them, we employ such a manner of supervision, and we can't expect much boost in the performance with this.

    \subsection{Unsupervised Co-skeletonization Results}
    The proposed method works in the unsupervised scenario too, where absolutely no annotations are provided. The proposed method under this scenario completely relies on the clustering process to retrieve suitable neighbors in a mixed image collection. Since SK-506 and SK-LARGE datasets are suitable for this purpose, we report our unsupervised co-skeletonization results in Table~\ref{tab:unsup} on these datasets while comparing with the two baselines. Note that these unsupervised results are on the entire dataset, not just the test part. In the unsupervised scenario as well proposed method outperforms the two baselines. Note that since SK-LARGE is a large dataset, we apply our key prior propagation approach.
    
    \begin{table}
        \begin{center}
            \begin{tabular}{|l|c|c|c|c|}
                \hline
                &SK-506&SK-LARGE \\\hline
                Ours$^0$ & 0.362&0.333\\
                Ours (w/o $in$)   & 0.365& 0.352\\    Ours&\textbf{0.475}&\textbf{0.429}\\
                \hline
            \end{tabular}
        \end{center}
        \vspace{1mm}
        \caption{Comparisons of the unsupervised co-skeletonization results of our method with the two baselines using $F^\alpha$ metric.}\label{tab:unsup}
    \end{table}

    \subsection{Original approach v/s Key Prior Propagation approach}
    The main difference between our original key prior propagation approach is the way our two priors, co-skeleton prior and co-segment prior, are generated. The latter requires a significantly lesser number of alignments compared to the first at the cost of an assumption that the alignments are precise. In Table~\ref{tab:apcomp}, we show how significant is the speed up and drop in the performance when both the approaches are applied to the WH-SYMMAX dataset. While there is a speed-up of three times, the drop in the performance is just marginal. Therefore, the assumption holds to a good degree.

    \begin{table}
        \begin{center}
            \begin{tabular}{|l|c|c|}
                \hline
                &    Run-time (in mins) & Performance \\
                \hline\hline
                Original &230&0.545\\
                Key Prior Propagation     &80& 0.532\\
                \hline
            \end{tabular}
        \end{center}
        \vspace{1mm}
        \caption{Comparison of our two approaches in terms of the run-time and the performance on WH-SYMMAX dataset. While there is a speed-up of almost three times, the drop in performance is marginal. Note that the time reported is the time taken for prior generation, not the entire time.}\label{tab:apcomp}
    \end{table}

    \subsection{Discussions and Limitations}
    \begin{figure}
        \begin{center}
            \includegraphics[width=1\linewidth]{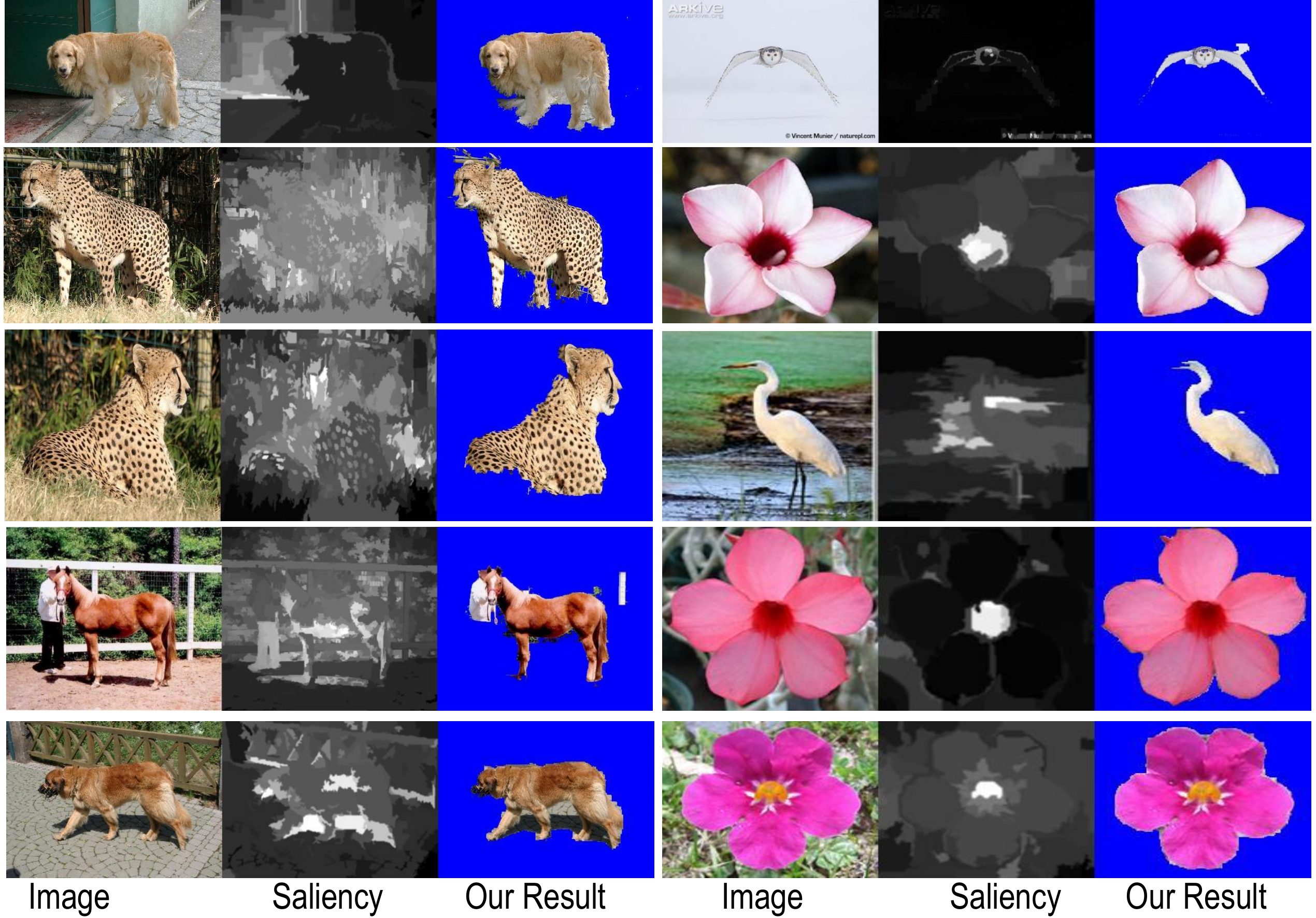}
        \end{center}
        \vspace{-4mm}
        \caption{Good segmentation examples despite the bad saliency maps}
        \label{fig:badint}
    \end{figure}
    We show few sampled segmentation results initialized by poor saliency map \cite{cheng2011global} in Fig.~\ref{fig:badint}. Despite such poor initializations, our algorithm manages to segment out the objects convincingly well, thanks to joint processing.
    Our method has some limitations. First, for initialization, our method requires common object parts to be salient in general across the neighboring images, if not in all. Therefore, it depends on the quality of the neighbor images. The second limitation lies in difficulty during the warping process. For example, when the neighboring images contain objects at different sizes or from different viewpoints, the warping processing finds it challenging to align the images well. However, such a situation is unlikely to occur when there are a large number of images, resulting in diversity to select appropriate neighbors. Another issue is that smoothing the skeleton may cause missing out some essential short branches. The third limitation occurs when a part occludes other parts of the object. As a result, the shape of the object doesn't look desirable for properly skeletonizing the objects. For example, a baseball player in Fig.~\ref{fig:coskelarge} misses out his left hand.
    

    \section{Conclusion}
    The major contributions of this paper lie in our novel object co-skeletonization problem and the proposed coupled co-skeletonization and co-segmentation framework, which effectively exploits inherent interdependencies between the two to assist each other synergistically. Extensive experiments demonstrate that the proposed method achieves very competitive results on different benchmark datasets, including our new CO-SKELARGE dataset, developed especially for weakly-supervised co-skeletonization benchmarking.

    \section*{Acknowledgment}
    This research is supported by the National Research Foundation, Prime Minister's Office, Singapore, under its IDM Futures Funding Initiative. It is also supported by the HCCS research grant at the ADSC\footnote{This work was partly done when Koteswar and Jiangbo were interning and working in ADSC} from Singapore's A*STAR.

	\ifCLASSOPTIONcaptionsoff
	\newpage
	\fi

	
	
	%
	\bibliographystyle{IEEEtran}
	\bibliography{main}
	
	%
	
	
	
	
	
	
	

\end{document}